\documentclass[10pt,twocolumn,letterpaper]{article}
\usepackage{authblk}
\usepackage{cvpr}              

\usepackage{graphicx}
\usepackage{amsmath}
\usepackage{amssymb}
\usepackage{booktabs}

\usepackage[dvipsnames]{xcolor}

\usepackage[pagebackref,breaklinks,colorlinks]{hyperref}

\usepackage[capitalize]{cleveref}
\crefname{section}{Sec.}{Secs.}
\Crefname{section}{Section}{Sections}
\Crefname{table}{Table}{Tables}
\crefname{table}{Tab.}{Tabs.}

\def\Vec#1{{\boldsymbol{#1}}}
\def\Mat#1{{\boldsymbol{#1}}}
\usepackage{algorithm}
\usepackage{algpseudocode}
\usepackage{booktabs}

\def\wacvPaperID{*****} 
\def\confName{WACV}
\def\confYear{2024}

\begin{document}

\title{SafeSea: Synthetic Data Generation for Adverse \& Low Probability Maritime Conditions}



\author[1]{Martin Tran}
\author[1]{Jordan Shipard}
\author[2]{Hermawan Mulyono}
\author[1,2]{Arnold Wiliem}
\author[1]{Clinton Fookes}

\affil[1]{\small Signal Processing, Artificial Intelligence and Vision Technologies (SAIVT), Queensland University of Technology, Australia}
\affil[2]{Sentient Vision Systems, Australia}
\affil[ ]{\textit {\tt \small \{minhtuan.tran@connect, jordan.shipard@hdr, c.fookes@\}qut.edu.au}, \textit{ \tt \small \{arnoldw, hermawanm\}@sentientvision.com}}

\maketitle
\begin{abstract}
    High-quality training data is essential for enhancing the robustness of object detection models. Within the maritime domain, obtaining a diverse real image dataset is particularly challenging due to the difficulty of capturing sea images with the presence of maritime objects , especially in stormy conditions. These challenges arise due to resource limitations, in addition to the unpredictable appearance of maritime objects. Nevertheless, acquiring data from stormy conditions is essential for training effective maritime detection models, particularly for search and rescue, where real-world conditions can be unpredictable. In this work, we introduce SafeSea, which is a stepping stone towards transforming actual sea images with various Sea State backgrounds while retaining maritime objects. Compared to existing generative methods such as Stable Diffusion Inpainting~\cite{stableDiffusion}, this approach reduces the time and effort required to create synthetic datasets for training maritime object detection models. The proposed method uses two automated filters to only pass generated images that meet the criteria. In particular, these filters will first classify the sea condition according to its Sea State level and then it will check whether the objects from the input image are still preserved. This method enabled the creation of the SafeSea dataset, offering diverse weather condition backgrounds to supplement the training of maritime models. Lastly, we observed that a maritime object detection model faced challenges in detecting objects in stormy sea backgrounds, emphasizing the impact of weather conditions on detection accuracy. The code, and dataset are available at \url{https://github.com/martin-3240/SafeSea}.
    

\end{abstract}    
\section{Introduction}
\begin{figure}
        \centering
        \includegraphics[scale=0.42]{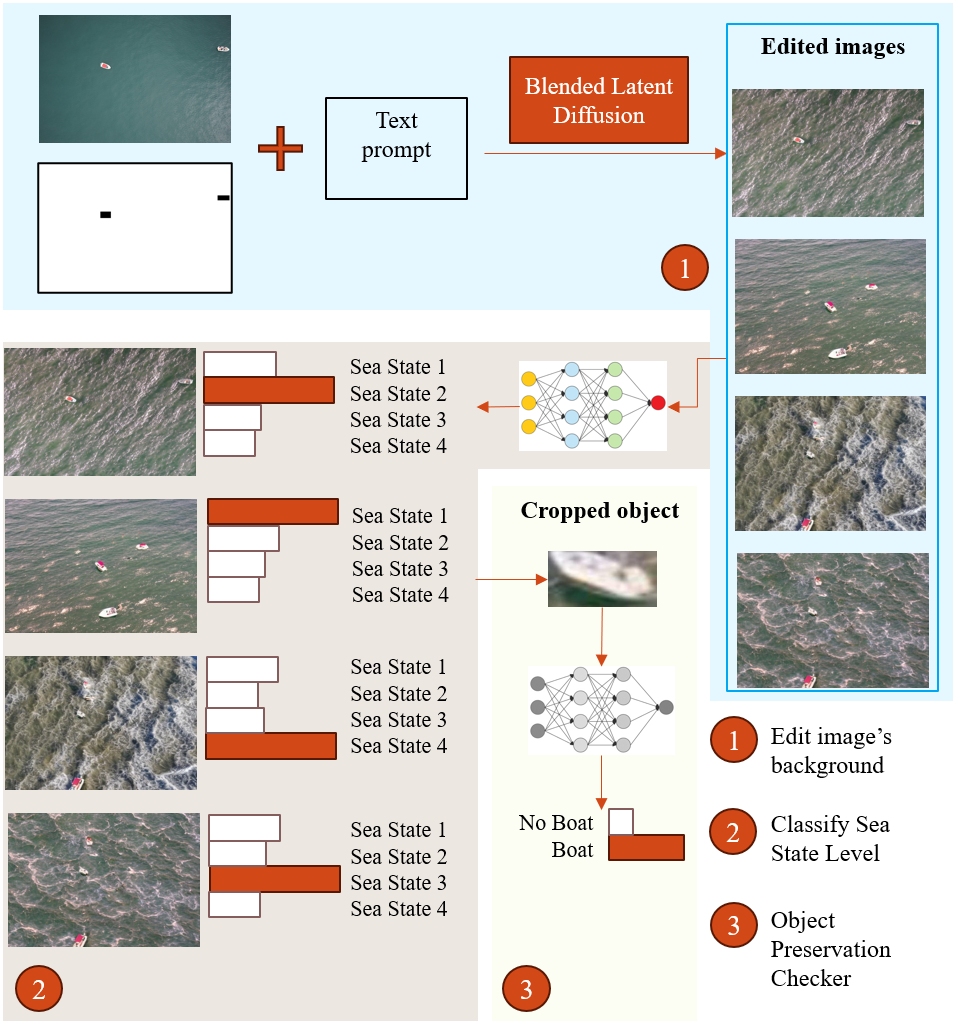}
        \caption{Diagram summarizing our method (SafeSea). Original maritime images' sea background is edited with a mask and text description using Blended Latent Diffusion \cite{BlendedLatentDiffusion}. The edited images are then classified into 4 Sea State categories before their marine objects (boats) are cropped according to the ground truth bounding box and checked for preservation.}
        \label{fig:method}
  \end{figure} 
  
 Climate change is increasing the likelihood of extreme weather events, such as large storms \cite{EPA_2023}, which can be destructive, especially for humans and boats in stormy waters. Quick and accurate rescue efforts are crucial in such situations as death can occur in as little as 50 minutes~\cite{Survival22}. Automated disaster response in the ocean is a growing area that uses advanced technology to help search and rescue in extreme weather events. Leveraging autonomous systems, remote sensing technologies, artificial intelligence, and real-time data analysis, automated disaster response aims to minimise risks, and accelerate response times. However, detecting entities in need of rescue during severe conditions in the ocean is often challenging. Therefore, there is a need for automated disaster responses that can operate effectively in these harsh conditions and can precisely locate objects, vessels, and people  requiring assistance. 
   
  Developing robust models to automatically detect objects in the sea during extreme weather conditions is a challenging task. The challenge primarily arises from the difficulty of obtaining high-quality data in such chaotic and unpredictable situations. Adverse weather conditions, particularly during storms, pose a safety risk, preventing the deployment of cameras in critical areas for data collection. Furthermore, the wild nature of disaster scenes often stretches already limited resources, including personnel, equipment, and communication channels, thus constraining the capacity for data acquisition \cite{sofar}. Due to the lack of datasets for training models to effectively detect maritime objects in the ocean during severe weather conditions, generating synthetic datasets to train and test object detection models can be highly beneficial. 
  
  By harnessing recent generative models such as Stable Diffusion (SD)~\cite{stableDiffusion}, and DALL-E~\cite{DALL-E}, it is now possible to generate realistic images based on textual descriptions. However, generating synthetic images solely based on textual descriptions experiences limitations in specifying object location, size, and type within the ocean scene, restricting the replication of realistic scenarios. To overcome this challenge, a more practical approach is to utilize real sea images with marine objects and performing transformations on the sea background. 
  
  Unfortunately, when applying the Stable Diffusion model to edit the real sea image's backgrounds, our experiments reveal a significant limitation in using text prompts to precisely control the desired Sea State. The edited background randomly appears not according to the description, so it would require checking manually to ensure the image's quality. Consequently, this process becomes time-consuming as it requires manual control to achieve a visually satisfactory Sea State. Furthermore, we found that existing image editing methods often replace or excessively modify the objects in the edited images, making the images unusable. Therefore, relying solely on synthetic image generation techniques to create a dataset with diverse Sea State Levels proves to be challenging, as it demands a substantial amount of time for generation and manual verification. 

  
  To tackle this issue, we propose a method for editing images of calm ocean scenes into stormy ocean scenes, focusing on UAV-view images. We replace the original calm ocean with an ocean environment corresponding to standard definitions from a Bureau of Meteorology as depicted in Table \ref{tab:SeaState}, all while attempting to retain the maritime objects in the images~\footnote{In this work we confine ourselves to only retaining the boats and reserving the other marine objects for future work.}.
  This method shows as a proof-of-concept to develop a simple-yet-effective approach, capable of automatically screening out poor-quality edited images with overly edited preserved objects. 

  Our work takes an input image and use an image generation method to perform background transformation.
  Recent work proposed the Blended Latent Diffusion ~\cite{BlendedLatentDiffusion} which demonstrates significantly improved results over the Stable Diffusion model in preserving foreground objects while also translating the background. Therefore, we employed Blended Latent Diffusion to modify images with object masking. Due to the stochastic nature of diffusion~\cite{diffusion2015}, we find that prompts alone can be unreliable for producing images according to the sea state definitions. Thus we apply the Sea State Classifier which classifies the transformed image into one of the sea state definitions. Then, the Object Preservation Checker is applied to evaluate whether the transformed image still preserve the objects from the input image.
    
  We leverage the SeaDroneSee dataset~\cite{varga2022seadronessee} for experimentation and validation. To quantitatively evaluate the effectiveness of our approach, we compare it against other image editing methods, such as Stable Diffusion Inpainting \cite{stableDiffusion}. Furthermore, we conducted object detection tests using YoloV5 model pre-trained on calm sea state images \cite{seadronesseeDetection} to assess the impact of various sea states on object recognisability in edited images. Our findings reveal that the pre-trained object detection model struggles to identify objects in increasingly stormy conditions. This suggests that the model's performance is less effective when it encounters previously unseen stormy sea backgrounds, making it more challenging to detect objects in rough sea surface conditions.

   \noindent \textbf{Contributions - } We list our contributions as follows: 
   \begin{enumerate}
    \item We propose a simple-yet-effective method for modifying the sea state level of the ocean in maritime images while preserving the objects of interest, enhancing their utility for various applications;
    \item We construct and propose a synthetic dataset, the SafeSea dataset, enabling the training of models to accommodate diverse weather conditions;
    \item We conduct an evaluation of the SafeSea dataset with YoloV5 to assess the model's performance under varying weather conditions.
   \end{enumerate}

We continue our paper as follows. Section~\ref{sec: relatedworks} presents an overview of prior work including maritime datasets and diffusion models. In Section~\ref{sec:problem}, we define our problem and the goal we want to achieve before Section~\ref{sec:method} outlines our proposed SafeSea method designed to accomplish this goal. The description of the SafeSea dataset are presented in Section~\ref{sec:dataset}. Section~\ref{sec:experiment} presents our experiment and discussion. We conclude the paper and discuss about limitations and future work in Sections~\ref{sec:conclusion} and Section~\ref{sec:futurework}.

\section{Related Works}
\label{sec: relatedworks}
\textbf{Sea Datasets - } Numerous public maritime datasets facilitate the training of marine object detection models. Notable examples include the VAIS dataset \cite{VAIS}, offering over 1,000 RGB and infrared image pairs, showcasing various ship types. The IPATCH dataset \cite{IPATCH} records realistic maritime piracy scenarios, while the SeaShips dataset \cite{SeaShips} features ship images from inland waterways. The 'Singapore Maritime Dataset' \cite{SIN} captures marine objects using onshore platforms and vessels, providing diverse perspectives. Additionally, the Seagull dataset \cite{Seagull} and Maritime SATellite Imagery (MASATI) dataset \cite{MASATI} offer aerial images. The SeaDronesSee dataset \cite{varga2022seadronessee} contains over 54,000 frames with around 400,000 instances. Despite their strengths, these datasets often focus on specific objects like ships and boats,  and often the objects of interest are relatively large compared to the image size in certain datasets. Furthermore, most of the images are taken under good weather conditions.

\noindent
\textbf{Synthetic Datasets - } Extensive research explores methodologies employing synthetic images for training object detection models. Noteworthy contributions include Peng et al. \cite{Peng} emphasizing the refinement of synthetic object backgrounds for improved detection reliability. The use of powerful game engines like Unity~\cite{unity} and Unreal Engine~\cite{unreal} is prominent, as demonstrated by Becktor et al. \cite{becktor} in the context of spacecraft guidance and control. Unreal Engine 4 ~\cite{unreal} has proven valuable in autonomous driving, Dosovitskiy et al. \cite{dosovitskiy}, and maritime image generation, as utilized by Becktor et al. \cite{Becktor1}. Kiefer et al. \cite{Kiefer} analyzed maritime and terrestrial images, incorporating real and synthetic data from the Grand Theft Auto V (GTAV) simulation platform \cite{Rockstar}. Xiaomin et al. \cite{xiaomin} introduced SeeDroneSim, utilizing the Blender game engine \cite{blender} for UAV-based maritime images. Airbus \cite{airbus} released a 2018 dataset of 40,000 satellite images designed for ship detection using synthetic aperture radar (SAR) technology. While existing datasets focus on synthetic objects of interest, our work concentrates on generating diverse environmental conditions based on real data.

\noindent
\textbf{Diffusion Models - } Diffusion models have been widely employed for image transformation in various contexts. Trabucco, Doherty et al. \cite{trabucco} employed pre-trained text-to-image diffusion models for semantic-based image modification. Shin et al. \cite{shin} utilized Stable Diffusion for image generation using the Textual Inversion \cite{textualInversion} method. Ron et al. proposed the Null-text Inversion method \cite{nulltext}, employing prompt-to-prompt text editing for image denoising. Rombach et al. \cite{stableDiffusion} introduced Stable Diffusion Inpainting for image inpainting using masks and a latent text-to-image diffusion model. Omri et al. \cite{BlendedLatentDiffusion} accelerated the transformation process with a lower-dimensional latent space. Our selection of the Blended Latent Diffusion approach \cite{BlendedLatentDiffusion} is based on its demonstrated superior results in editing image backgrounds.

Presently available real-image maritime datasets have limitations, including fixed camera positions and a restricted variety of marine objects. Furthermore, they often lack diversity in representing various weather conditions reflecting on the sea background. This lack of diversity can restrain the development of a high-quality dataset for training deep-learning models. To address this challenge, numerous studies have been conducted to generate high-quality synthetic images capable of replicating real-world scenarios. These synthetic images can be produced by leveraging game engines such as Unity \cite{unity} and Unreal Engine \cite{unreal} or by employing diffusion models to modify real images\cite{trabucco, shin, textualInversion, hertz2023prompttoprompt, stableDiffusion, BlendedLatentDiffusion}. While using game engines can be resource-intensive, editing images with diffusion models offers a simpler approach. Unfortunately, the editing process via diffusion models is still time consuming and labour intensive due to imprecise control that specifies which part of the image that needs to be edited.
This works proposes a proof-of-concept to enable automation in the editing process which significantly reduces the time and labour whilst maintaining the quality of the edited images.
\begin{table}[]
\begin{tabular}{cll}
Sea State & Definition\\ \hline\hline
1         &  \vspace{3pt} \parbox{6.2cm}{The water exhibits a gentle ripple, devoid of breaking waves, featuring a low swell of short to average length occasionally.}               \\  \hline
2         &   \vspace{3pt} \parbox{6.2cm}{Slight waves breaking, with smooth waves on the water surface}        \\ \hline
3         &  \vspace{3pt}  \parbox{6.2cm}{Mildly increased waves, leading to some rock buoys and causing minor disturbances for small craft}              \\ \hline
4         &  \vspace{3pt} \parbox{6.2cm}{The sea takes on a furrowed appearance, characterized by moderate waves}             \\ \hline
\end{tabular}
\caption{Sea States descriptions as defined by the ABS}
\label{tab:SeaState}
\end{table}

\section{Problem Definition}
\label{sec:problem}
\begin{figure}[t]
\centering
  \begin{minipage}[b]{0.23\textwidth}
    \includegraphics[width=\textwidth]{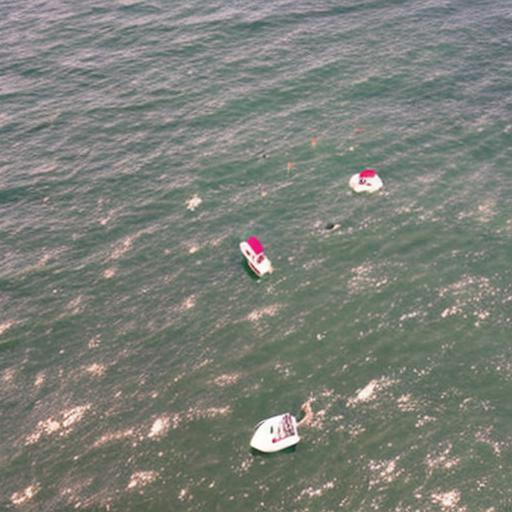}
    \caption*{Sea State Level 1}
  \end{minipage}
  \hfill
    \begin{minipage}[b]{0.23\textwidth}
    \includegraphics[width=\textwidth]{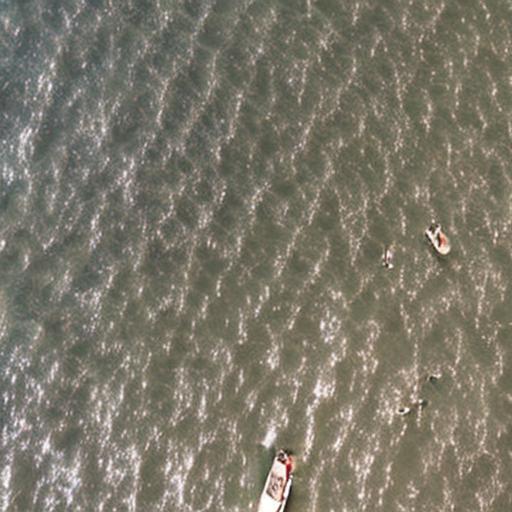}
    \caption*{Sea State Level 2}
  \end{minipage}
  \begin{minipage}[b]{0.23\textwidth}
    \includegraphics[width=\textwidth]{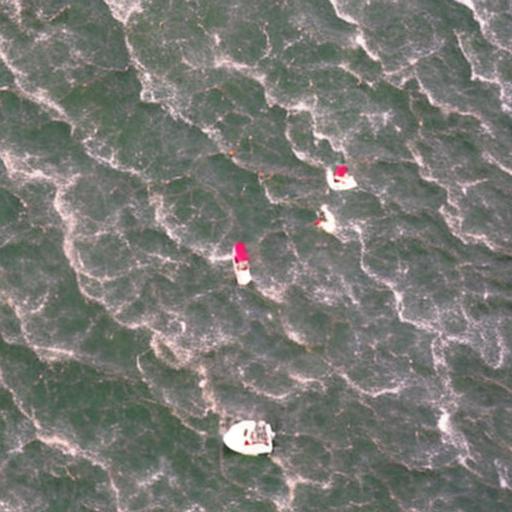}
    \caption*{Sea State Level 3}
  \end{minipage}
  \hfill
\begin{minipage}[b]{0.23\textwidth}
    \includegraphics[width=\textwidth]{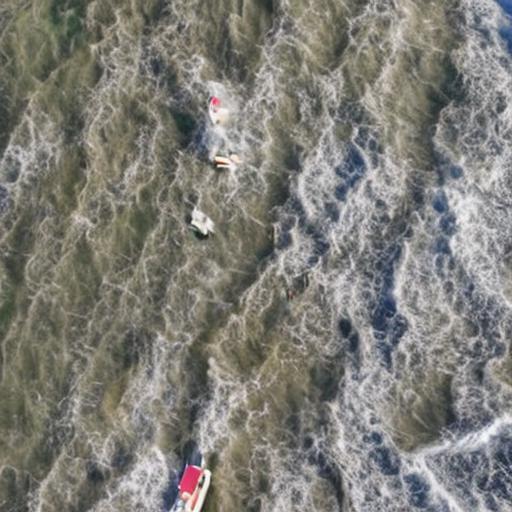}
    \caption*{Sea State Level 4}
  \end{minipage}
\caption{Examples of edited images belong to different classified sea state levels. The Sea State level information is provided to these images after the editing process. The sea surface becomes more dynamic with waves and whitecaps as the Sea State Level increases.}
\label{fig:SS}
\end{figure}

Here we technically define our problem. Starting from an image $\Mat{X}$ containing a set of objects, $\mathcal{B} = \{ \Vec{b}_i \}^{N}_{i=1}$, where $N$ is the number of objects, and $\Vec{b}_i$ is the $i$-th object bounding box. 
We first define the foreground $f_x$ of image $\Mat{X}$ as all areas of $\mathcal{B}$, subsequently, we define the background $bg_x$ as the inverse of all areas of $\mathcal{B}$ in $\Mat{X}$. Initially, $bg_x$ is a calm ocean environment. We wish to replace $bg_x$ such that it corresponds to a particular sea state, $SS$, where $SS$ ranges from 1 to 4 and are defined in Table~\ref{tab:SeaState}, with examples shown in Figure~\ref{fig:SS}. We only aim to replace $bg_x$, and aim to retain the number of $N$ as was present in $\Mat{X}$. We refer to the resulting image as $\Mat{Y}$. Specifically, let $f_y$ and $bg_y$ as the foreground, and the background of $\Mat{Y}$, respectively. The goal is to have $bg_y \neq bg_x$, and $f_y \approx f_x$. 


\section{Proposed SafeSea method}
\label{sec:method}


The overall diagram of the propose method is depicted in Figure~\ref{fig:method} and the pseudo code is presented in Algorithm~\ref{alg:agl}. The method has three main components: (1) the image generation module which transform the background of an input image; (2) Sea State Level classifier; and (3) the Object Preservation Checker. Once an image is generated, its quality is automatically assessed by using (2) and (3). Specifically, the Sea State Level classifier determines the sea state level of the generated image, and the Object Preservation Checker ensures the generated image still contains the objects from the input image. 

The following section first discusses the image generation module which is powered by a diffusion model. Then, the Sea State Level classifier and the Object Preservation Checker modules are presented. 


\begin{algorithm}
[t]\captionsetup{labelfont={sc,bf}, labelsep=newline}
  \caption{Pseudocode for generating synthetic images of the sea with marine objects from real images as proposed in SafeSea method. The real image has its sea background edited using Blended Latent Diffusion \cite{BlendedLatentDiffusion}, then its Sea State level is provided by the Sea State Classifier. Finally, the original objects are checked if they are retained in the edited image, which lead to the decision of saving the image if at least one object is preserved.}
  \label{alg:agl}
\hspace*{\algorithmicindent} \textbf{Input:} Real sea images with marine objects ($Rs$), matching mask images ($Ms$) with ground true bounding boxes of marine objects are masked as black, text description of background ($P$)\\
\hspace*{\algorithmicindent} \textbf{Output:} Transformed images with edited background reflected different sea condition and the marine objects are preserved
\begin{algorithmic}[1]
\For{each $X$ in $Rs$} 
    \State $M \gets$ Corresponding mask image from $Ms$
    \State $Y\gets$ Blended\_Latent\_Diffusion($X, M, P$) \Comment{Edited image} 
    \State $SS_E\gets$ Sea\_State\_Classifier($Y$) \Comment{Find the edited image's Sea State}
    \State $E_R\gets$ $Y$ \Comment{Resize the edited image}
    \State $Cs\gets$ $E_R$ \Comment{Crop objects from the edited image}
    \For{each $C$ in $Cs$}
    \If{Object\_Preservation\_Checker($C$) = boat}
        \State Save $Y$
        \\        \hspace*{\algorithmicindent}\hspace*{\algorithmicindent}\hspace*{\algorithmicindent}\textbf{break}
    \EndIf
    \EndFor
\EndFor
\end{algorithmic}
\end{algorithm}

\subsection{Image Generation Module}

Diffusion models~\cite{ho2020denoising} are capable of being trained with guiding input channels, enabling them to perform conditional image generation such as creating synthesizing visual content based on textual descriptions~\cite{dalle}. Beyond synthesis, diffusion models are versatile image editors, allowing for targeted modifications. In this process, noise is added to the original image and subsequently denoised in response to an optional prompt, which describes the new image. Using masking techniques, these models can carry out semantic editing, selectively altering specific regions within the images while leaving others intact. Various approaches exist for utilizing Stable Diffusion in image editing, which can be broadly categorized into two main groups: those that employ pre-specified masks and those that mask images based on provided text descriptions.

In this work, we applied Blended Latent Diffusion method \cite{BlendedLatentDiffusion} to edit original sea images.
Blended Latent Diffusion provides a versatile approach to image editing, utilizing masks alongside guided text prompts. In a nutshell, it provides a versatile approach to image editing, utilizing masks alongside guided text prompts. To alter an image, a mask of identical dimension to that image is applied to designate the region to be modified, while guided text instructions define how the edited area should appear. The intended outcome is an image in which the masked region undergoes alterations while the unmasked portion remains unaltered. In the transformation process, the original image is encoded into a latent space, introducing a controlled level of noise. Additionally, the mask is downsampled within this latent space. At each stage of this process, the noisy latent image is subjected to denoising and is regarded as the foreground to be blended with the background elements. We observe that the method demonstrates promise in generating images with different ocean backgrounds whilst maintaining a considerable level of object preservation, in which the object can be recognized visually after the background edition process.


\subsection{Sea State Classifier}
To provide information about Sea State level in the edited image's background, we employed a Sea State classifier. We selected four distinct Sea States for our study, namely Sea States 1, 2, 3, and 4, as these are the only Sea States for which datasets are publicly available. The Sea State definitions are shown in Table \ref{tab:SeaState} .It is important to note that without recorded weather conditions at the time of image capture, visually classifying sea images into different Sea States is a challenging task. Leveraging the Manzoor-Umair Sea State Image Dataset (MU-SSiD) \cite{ssclassifier}, we trained a DenseNet~\cite{densenet} to categorize images into the four Sea State categories. The trained model achieved an accuracy rate of 71\% against the testing dataset.

\subsection{Object Preservation}
To evaluate whether an object remains preserved following image transformation, we developed an object preservation checker. This checker's primary function is to identify objects that are no longer recognizable within the provided ground truth bounding box. To do this, we train a binary classifier on a dataset consiting of two classes: "boat" and "not boat." The "boat" class contains images of cropped boats sourced from the ground truth SeaDroneSee dataset~\cite{varga2022seadronessee}, complemented by their augmented versions, which include flipping and blurring. In contrast, the 'not boat' class comprises images from the negative class extracted from the Boat-MNIST dataset \cite{varga2022seadronessee}, as well as crops from randomly selected backgrounds from synthetic images. Additionally, it includes crops intentionally containing small portions of boat objects from the edited images.These crops are generated by following the object ground truth bounding box, ensuring that they contain only one-fourth of the bounding box's area with the rest outside. Using the dataset with only horizontal flip data augmentation for the boat class. We trained a DenseNet \cite{densenet} model with a training batch size of 32, a fixed learning rate of 1e-5 without decay, and utilized the Adam optimization algorithm during the training process. Overall, the model achieves the accuracy of 74.86\% against testing dataset, including crops of boat objects in real images and non-boat crops from real and edited images. 

\begin{figure}[t]
    \centering
      \begin{minipage}[b]{0.23\textwidth}
        \includegraphics[width=\textwidth]{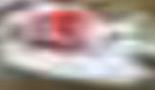}
        \caption*{}
      \end{minipage}
      \hfill
        \begin{minipage}[b]{0.23\textwidth}
        \includegraphics[width=\textwidth]{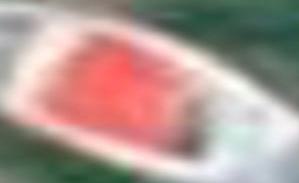}
        \caption*{}
      \end{minipage}
      \begin{minipage}[b]{0.23\textwidth}
        \includegraphics[width=\textwidth]{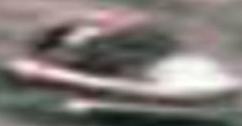}
        \caption*{}
      \end{minipage}
      \hfill
    \begin{minipage}[b]{0.23\textwidth}
        \includegraphics[width=\textwidth]{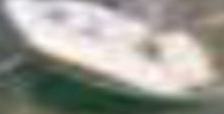}
        \caption*{}
      \end{minipage}
    \caption{Examples of boat object crops from edited images. The crops are taken based on the ground truth bounding boxes provided from the source images in the 'SeaDroneSea' dataset \cite{varga2022seadronessee}.}
    \label{fig:crop}
\end{figure}

Subsequently, based on the given ground truth bounding box information, we extracted boat objects from the edited images for evaluation using the trained checker. We conducted random visual checks on the preserved boat objects to evaluate the model's performance. While there are occasional misclassifications of objects as boats that do not visually resemble boats, the model generally achieves satisfactory accuracy in detecting non-boat objects, in which it can pick up cropped objects that resemble boats and filter out non-boat crops according to the visual checks with the accuracy of 69.45\%. Examples of the cropped boat objects are illustrated in Figure \ref{fig:crop}.

\section{SafeSea dataset}
\label{sec:dataset}

\begin{figure}[t]
\centering
  \begin{minipage}[b]{0.23\textwidth}
    \includegraphics[width=\textwidth]{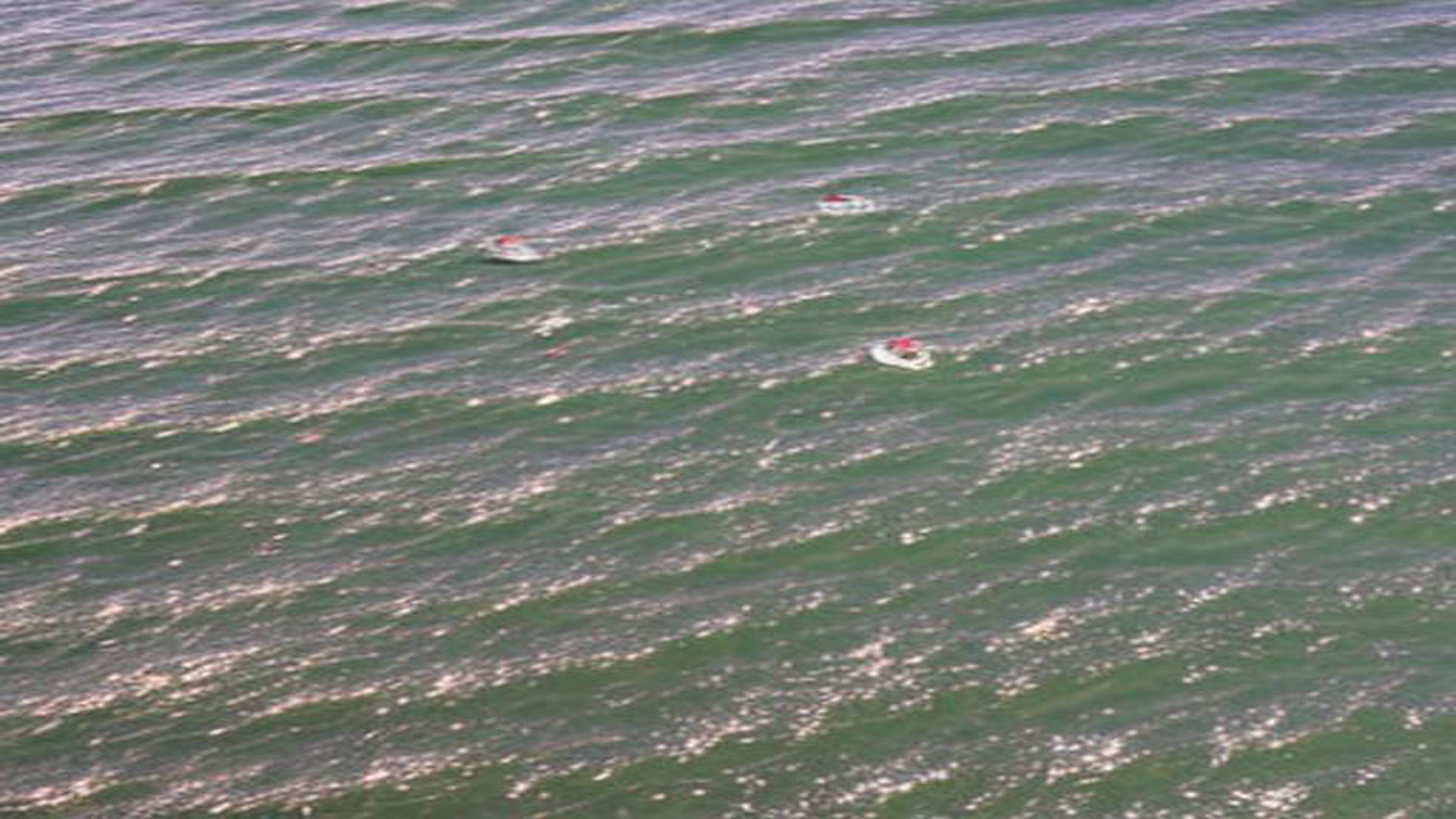}
    \caption*{Sea State Level 1}
  \end{minipage}
  \hfill
    \begin{minipage}[b]{0.23\textwidth}
    \includegraphics[width=\textwidth]{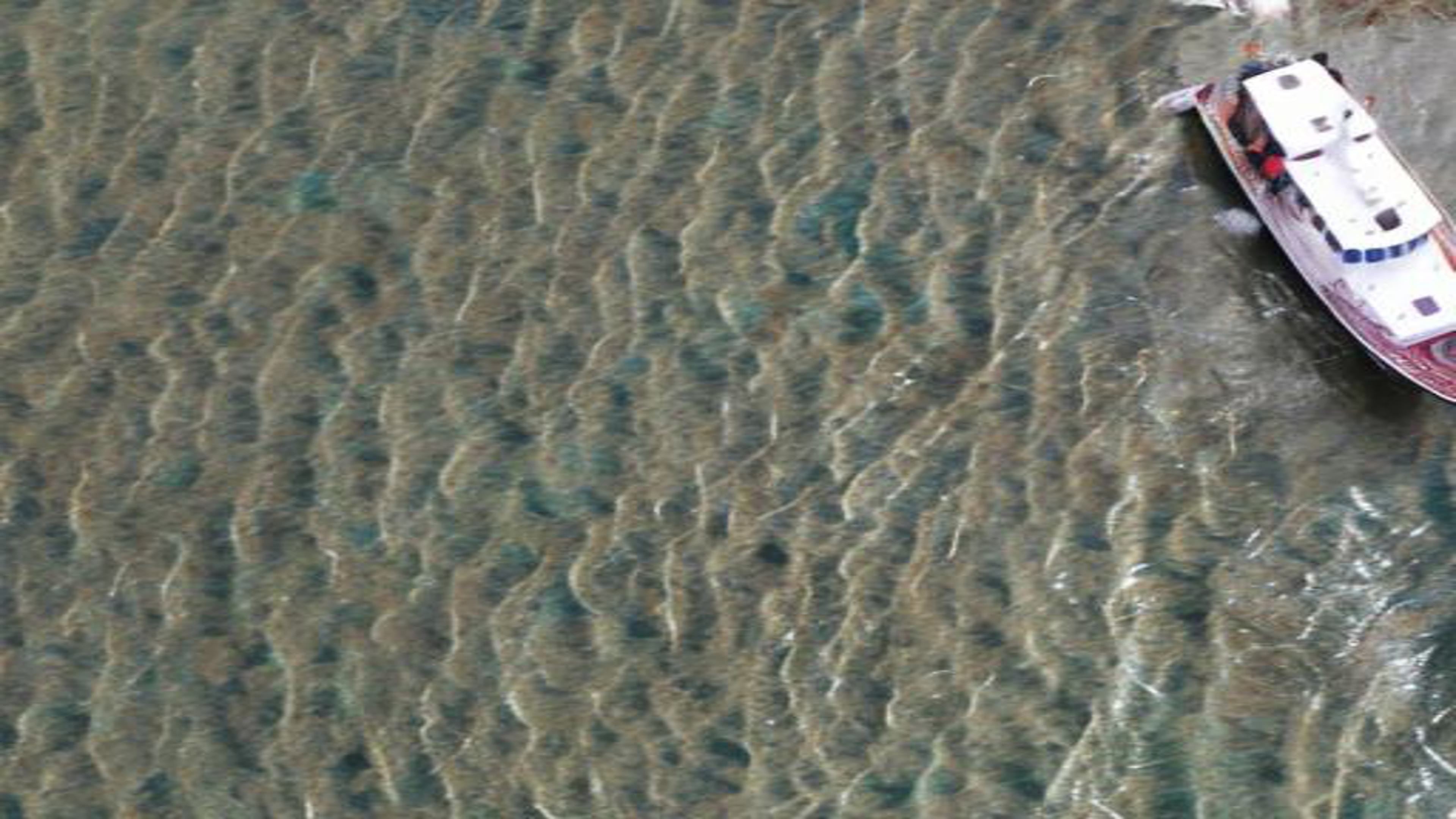}
    \caption*{Sea State Level 2}
  \end{minipage}
  \begin{minipage}[b]{0.23\textwidth}
    \includegraphics[width=\textwidth]{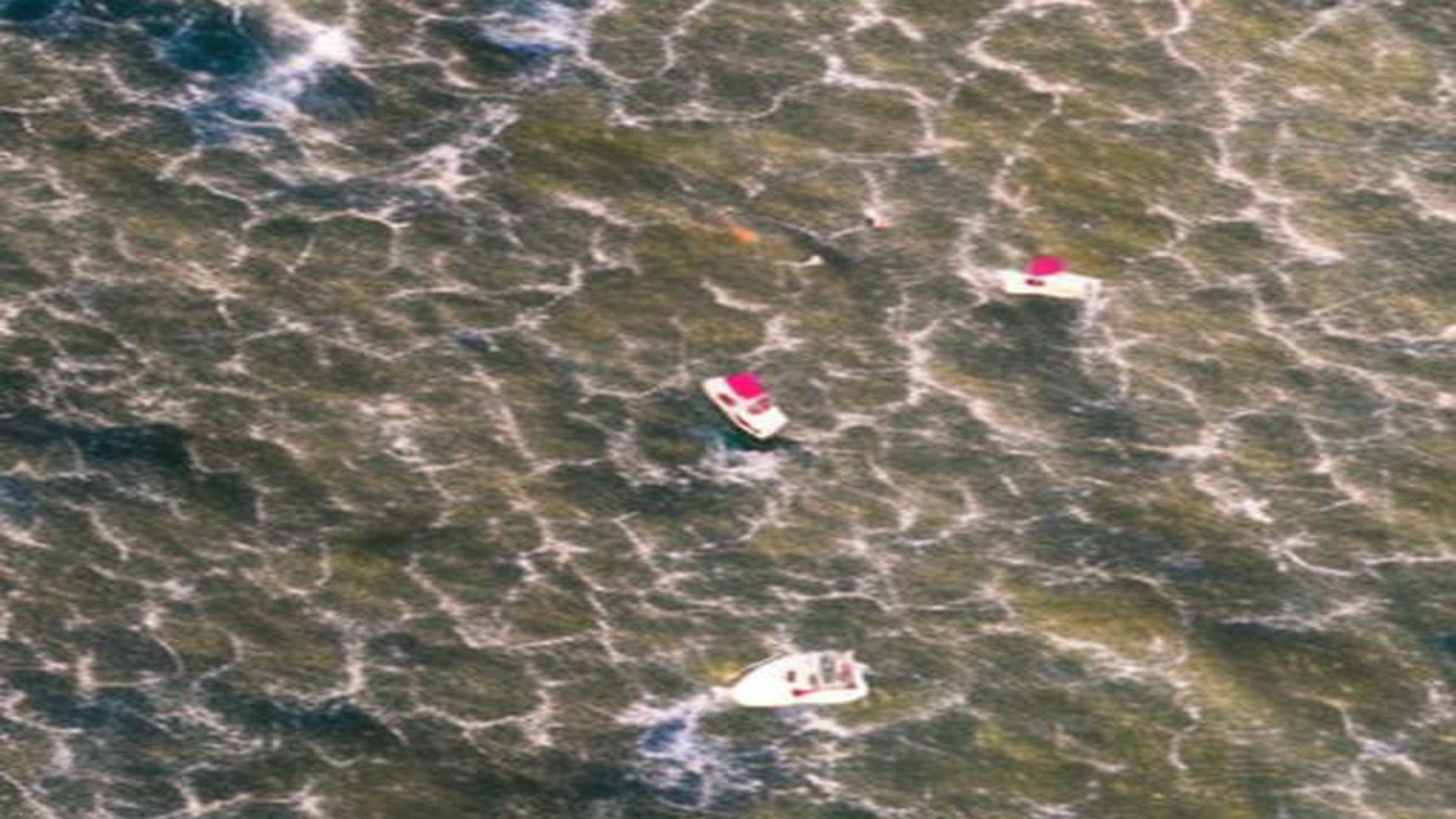}
    \caption*{Sea State Level 3}
  \end{minipage}
  \hfill
\begin{minipage}[b]{0.23\textwidth}
    \includegraphics[width=\textwidth]{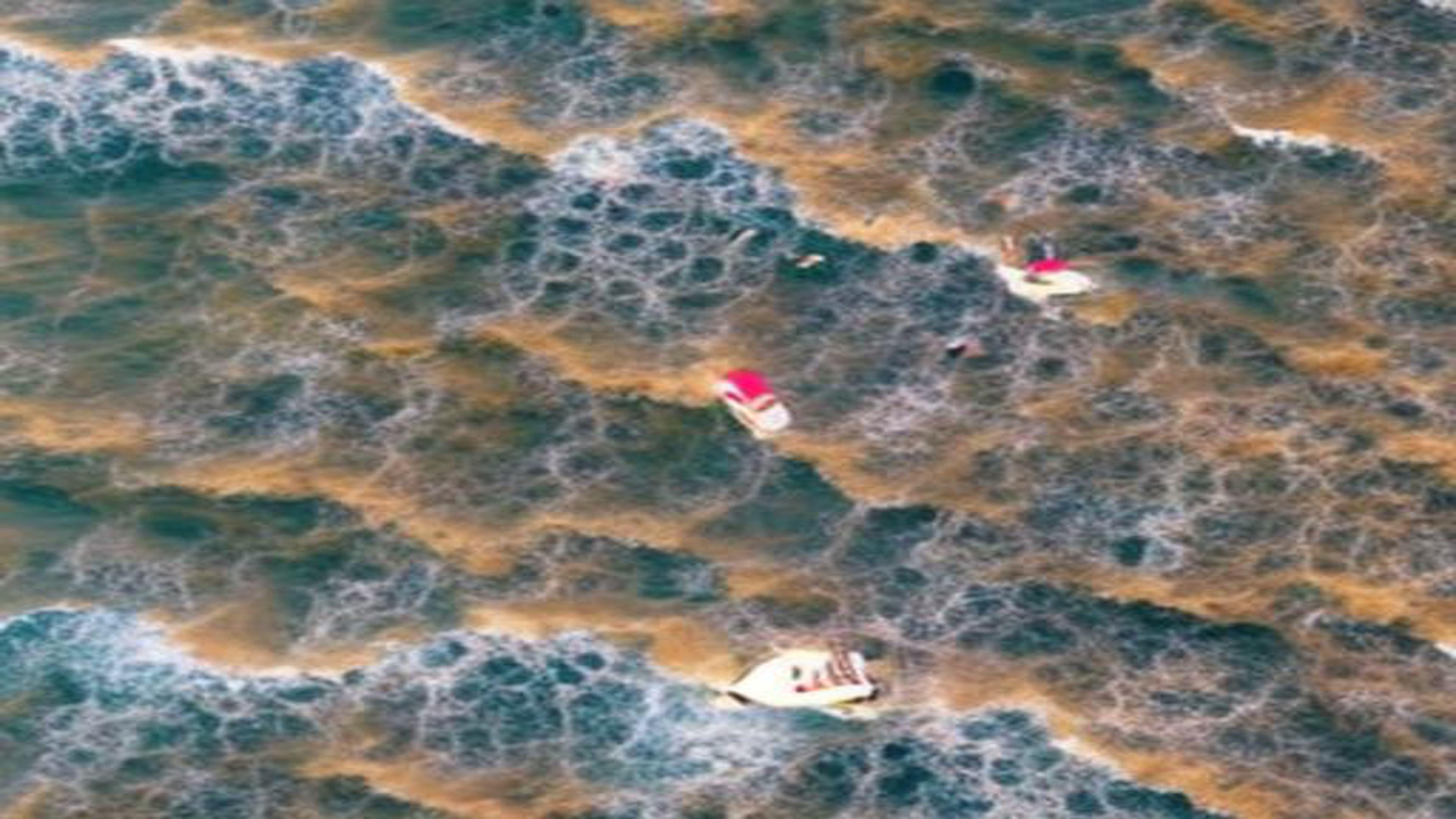}
    \caption*{Sea State Level 4}
  \end{minipage}
\caption{Examples of images in the SafeSea dataset. There is one example representing each Sea State level in the dataset.}
\label{fig:PS}
\end{figure}

The SafeSea dataset is created using the SafeSea method, involving the transformation of 300 calm ocean background images originally sourced from the 'SeaDroneSee' dataset \cite{varga2022seadronessee}. All edited images were resized to match the dimensions of their respective originals. 
The SafeSea method produces 69,694 images images. These are then classified into one of the sea state levels. The distribution of these images across different sea state levels is detailed in Table \ref{table:1}.
Notably, Sea State levels 3 and 1 exhibit the most and the least number of images, boasting 45,066 and 2,087 images respectively. In the intermediate Sea State 2 category, there are 19,390 images, while Sea State 4 includes 3,151 images. For a visual representation, a selection of dataset examples is provided in Figure~\ref{fig:PS}.

\vspace{-1mm}
\section{Experiment}
\label{sec:experiment}

This section is divided into two parts. The first part evaluates the proposed SafeSea method efficacy in filtering compared with the other methods such as the vanilla Stable Diffusion Inpainting. Then, we study the performance of YoloV5 detection model on the proposed SafeSea dataset.
\subsection{SafeSea method evaluation}
We first describe the experiment setup and then present the results afterwards.
\subsubsection{Experiment setup}

\noindent
\textbf{Evaluation protocol - } We generate 100 images from each evaluated method. Then, each image is manually checked by humans and categorized into good quality of bad quality group. A good quality image is defined as the following rules:
\begin{itemize}
    \item The background should contain either island, ocean or cloud. For instance if the background contains unexpected objects such as a fridge, then the image is deemed low quality.
    \item The background should look realistic
    \item All the objects should look visually acceptable. At least one boat is preserved in the image.
\end{itemize}
Each method is then compared by looking at its percentage of generating good quality images.

\noindent
\textbf{Methods - } The proposed SafeSea method is compared against two baselines: (1) the vanilla Stable Diffusion inpainting~\cite{stableDiffusion}; and (2) the vanilla Blended Latent Diffusion~\cite{BlendedLatentDiffusion}. Details for each baseline parameters is presented as follow.

\noindent
\textbf{Stable Diffusion inpainting (SD Inpainting) - } We use Stable Diffusion v1-4~\cite{stableDiffusion} with batch size of one. The masked image is derived from the groundtruth bounding boxes. The method is then fed with the following prompts to generate the images.
\begin{itemize}
    \item Sea State 1: Aerial image of the sea's surface. The water is gently rippled with no waves breaking. Canon EOS R3, Nikon d850 400mm, Canon DSLR, lens 300mm, 4K, HD.
    \item Sea State 2: Aerial image of the sea's surface. There are slight waves breaking with smooth wave on surface. Canon EOS R3, Nikon d850 400mm, Canon DSLR, lens 300mm, 4K, HD.
    \item Sea State 3: Aerial image of the sea's surface. Mild Waves are slight causing rock buoys and small craft. Canon EOS R3, Nikon d850 400mm, Canon DSLR, lens 300mm, 4K, HD.
    \item Sea State 4: Aerial image of the sea's surface. The water has furrowed appearance with moderate waves. Canon EOS R3, Nikon d850 400mm, Canon DSLR, lens 300mm, 4K, HD.        
\end{itemize}

\noindent
\textbf{Blended Latent Diffusion (BLD) - } 
We use Stable Diffusion v2-1-base \cite{stableDiffusion} with batch size of ten. Similar to the method above, we use the derived mask image from ground truth bounding boxes. We observe that when editing the sea background of an image to various Sea State Level, the reliance on prompts alone is insufficient. Therefore, to simplify the generation issue with prompt we only use one prompt in our experiments, which is ``Aerial image of sea's surface. Canon EOS R3, Nikon d850 400mm, Canon DSLR, lens 300mm, 4K, HD". The prompt allows generation of images with varying sea state levels with different generation seeds. 


\noindent
\textbf{SafeSea (proposed) - } We utilize Blended Latent Diffusion with the same parameters as above to edit original images. Then we use the sea state level classifier to determine the sea state level of the generated images. The Object Preservation Checker ensures objects are preserved. The generated image will be preserved if it preserves at least one object. 
Same as the other methods, the mask image is derived from the groundtruth bounding boxes provided by the SeaDroneSee dataset~\cite{seadronesseeDetection}. Note that this dataset contains several classes, but in this experiment we only aim to preserve the boat as boats have much larger object size.



\subsubsection{Results}

Table~\ref{table:manualCheck} presents the comparison results. The results suggest that the proposed SafeSea outperforms the baselines. As expected the BLD produces better image quality than the SD Inpainting as also shown in~\cite{BlendedLatentDiffusion}

It is clear that SafeSea outperforms the other methods, as it generates more good images with a realistic sea background and maintains object integrity more effectively.

\begin{table}[h!]
\centering
\begin{tabular}{@{}lcc@{}}
    \toprule
     & Good images (in \%)
      \\ 
     \midrule
     SD Inpainting~\cite{stableDiffusion}  & 26.5\% $\pm$ 5.94 \\
     BLD~\cite{BlendedLatentDiffusion}  & 52.94\% $\pm$ 8.25 \\
     SafeSea (proposed) & \textbf{63.59\%} $\pm$ 2.76 \\ 
     \bottomrule
\end{tabular}
 \caption{Comparison between SD Inpainting, BLD and SafeSea when we manually check quality of the 100 generated images from each method. The good image rate computes the percentage of good images passed by manually check by humans. High good image rate suggests the method works better for the task.}
 \label{table:manualCheck}
 \end{table}

\begin{figure}
\centering
  \begin{minipage}[b]{0.23\textwidth}
    \includegraphics[width=\textwidth]{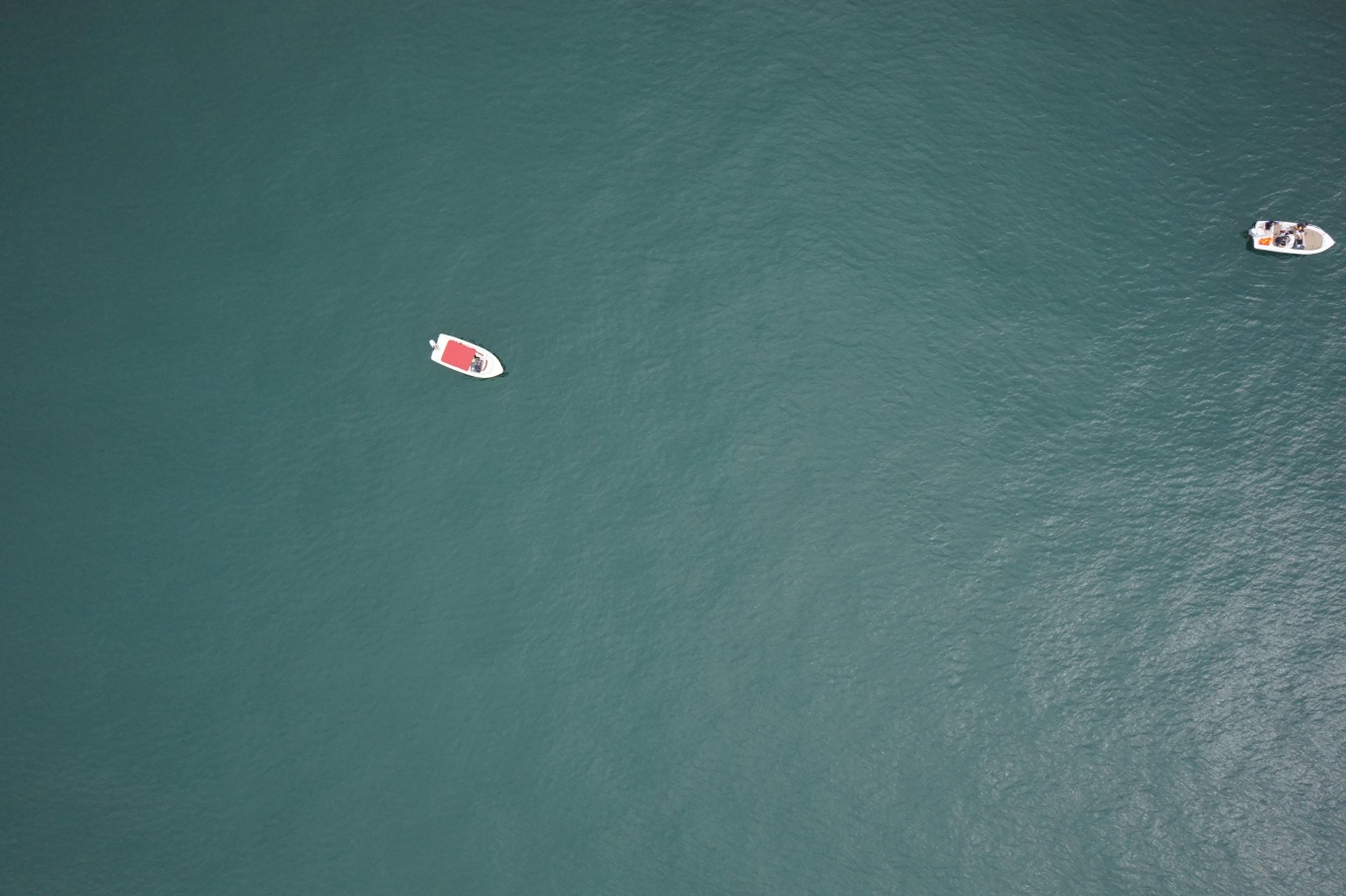}
    \caption*{Original}
  \end{minipage}
  \hfill
    \begin{minipage}[b]{0.23\textwidth}
    \includegraphics[width=\textwidth]{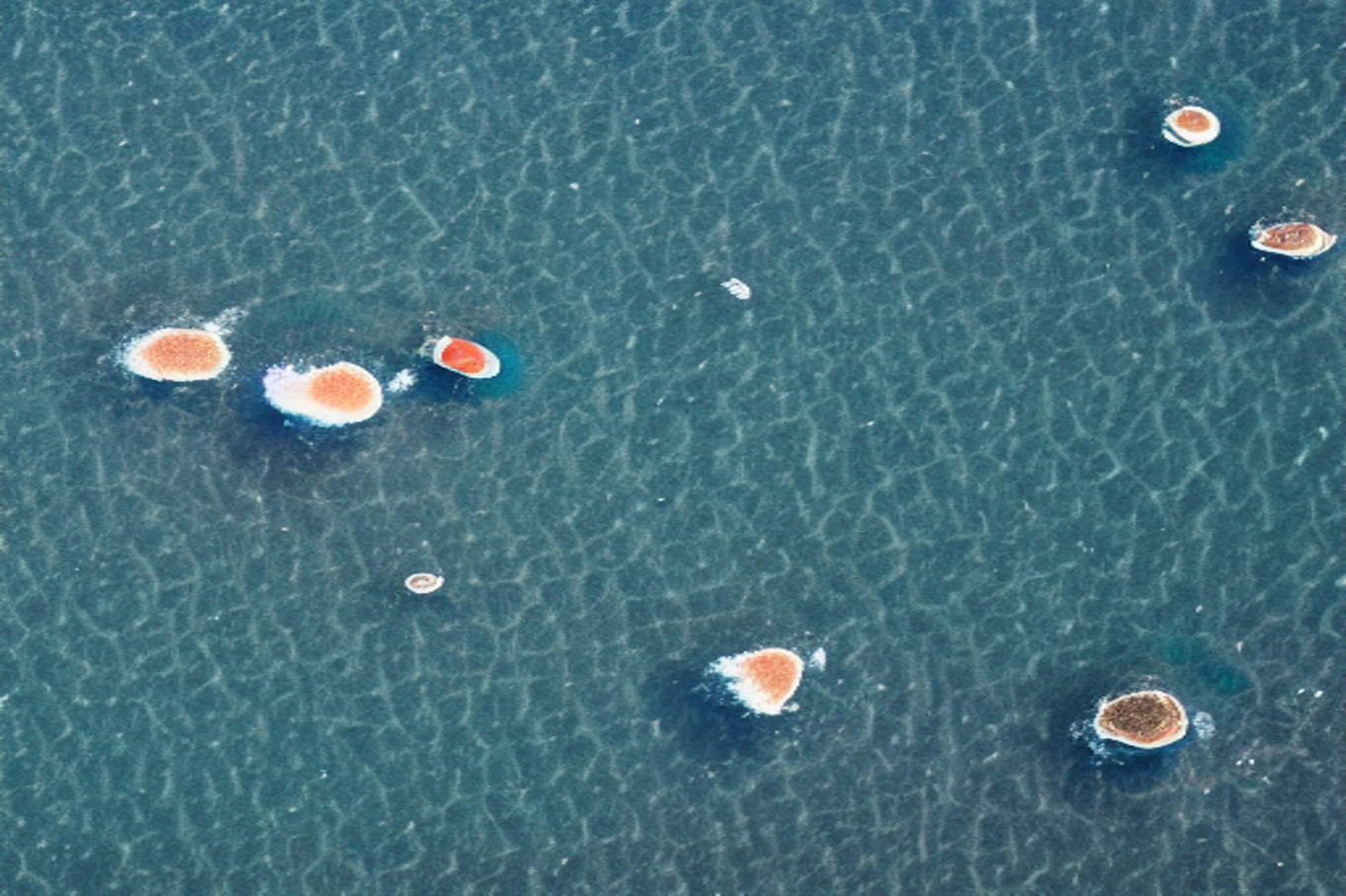}
    \caption*{Edited}
  \end{minipage}
  \begin{minipage}[b]{0.23\textwidth}
    \includegraphics[width=\textwidth]{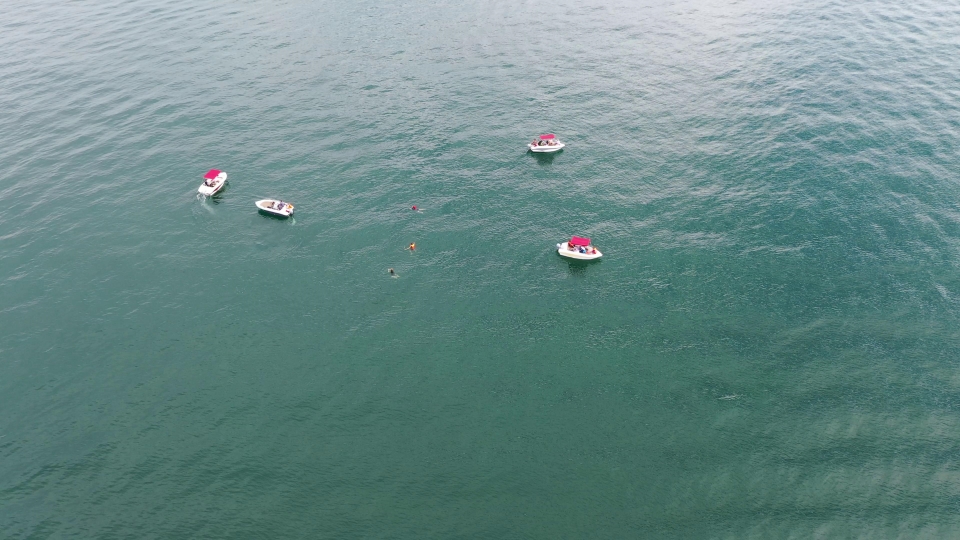}
    \caption*{Original}
  \end{minipage}
  \hfill
\begin{minipage}[b]{0.23\textwidth}
    \includegraphics[width=\textwidth]{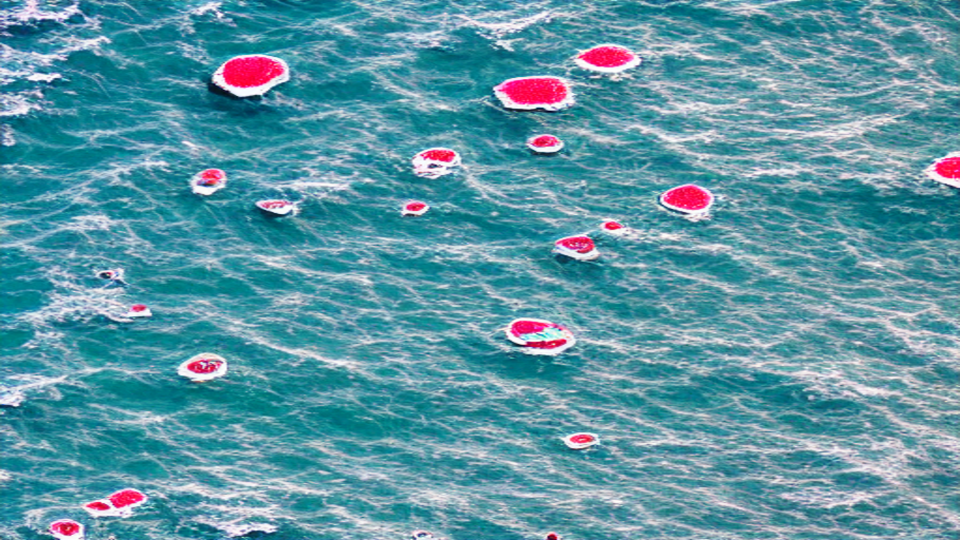}
    \caption*{Edited}
  \end{minipage}
  \begin{minipage}[b]{0.23\textwidth}
    \includegraphics[width=\textwidth]{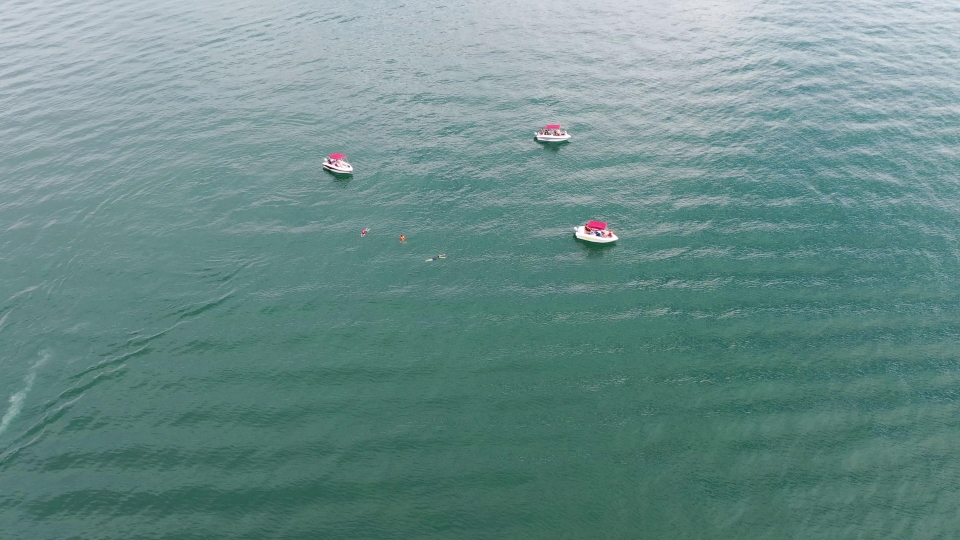}
    \caption*{Original}
  \end{minipage}
  \hfill
\begin{minipage}[b]{0.23\textwidth}
    \includegraphics[width=\textwidth]{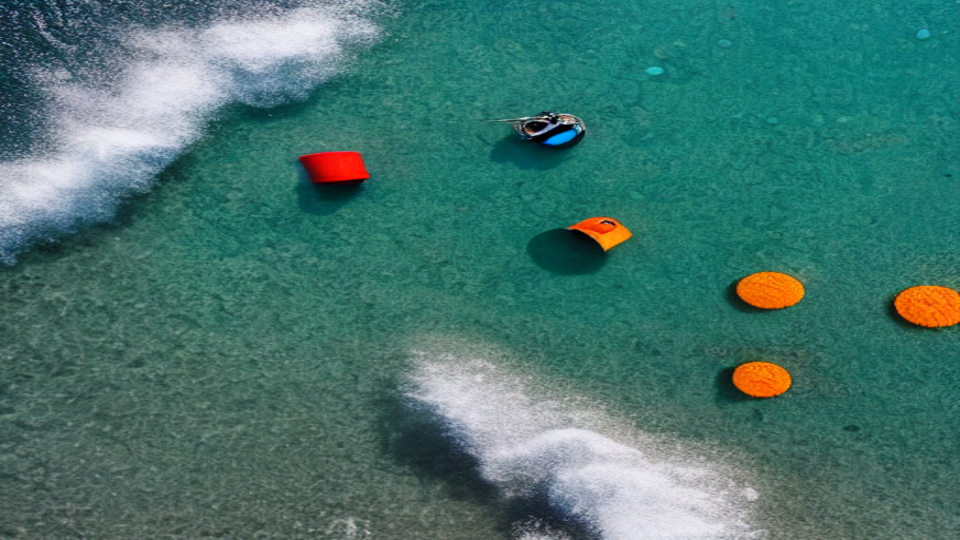}
    \caption*{Edited}
  \end{minipage}
\caption{Failure edited images using SD Inpainting. It suggests that this method can edit the image's sea background, however, the objects are not retained well, and several irrelevant objects are introduced unintentionally.}
\vspace{-5mm}
\label{fig:inpainting}
\end{figure}

While the result of Stable Diffusion Inpainting appear to achieve realism, many objects are blended into the background, or overly edited. Additionally, there are other inherent issues, such as other objects being inserted into the edited image. It appears as though the objects of interest may be inadvertently duplicated and dispersed into the background, as shown in figure \ref{fig:inpainting}. Although some edited background is of considerable quality, many of objects of interest are not retained. Additionally, the edited images may not be suitable for use due to the introduction of irrelevant modified objects and unexpected background.

\begin{figure}[t]
\centering
  \begin{minipage}[b]{0.23\textwidth}
    \includegraphics[width=\textwidth]{sections/images/original/1309.jpg}
    \caption*{Original}
  \end{minipage}
  \hfill
    \begin{minipage}[b]{0.23\textwidth}
    \includegraphics[width=\textwidth]{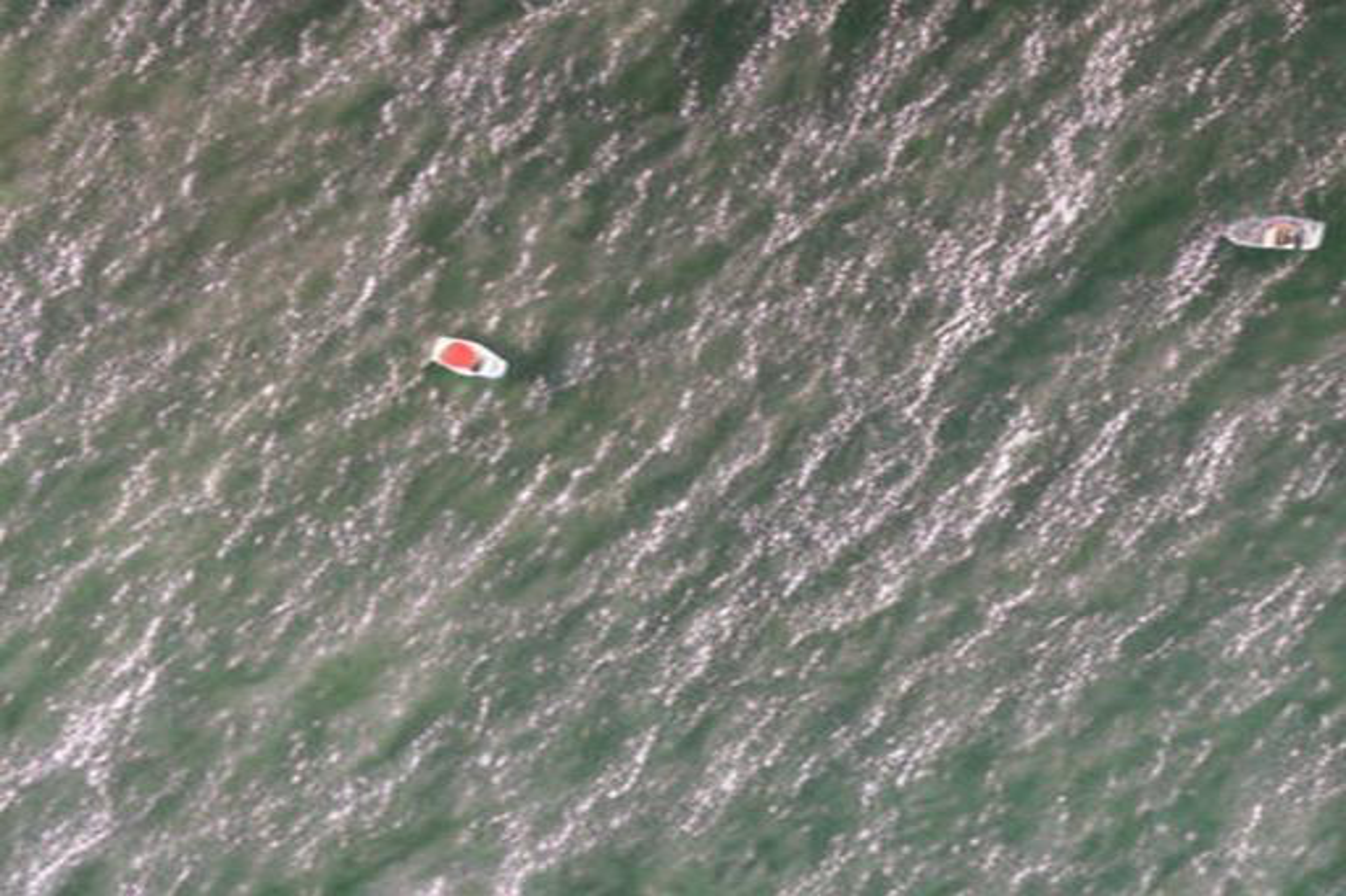}
    \caption*{Edited}
  \end{minipage}
  \begin{minipage}[b]{0.23\textwidth}
    \includegraphics[width=\textwidth]{sections/images/original/10774.jpg}
    \caption*{Original}
  \end{minipage}
  \hfill
\begin{minipage}[b]{0.23\textwidth}
    \includegraphics[width=\textwidth]{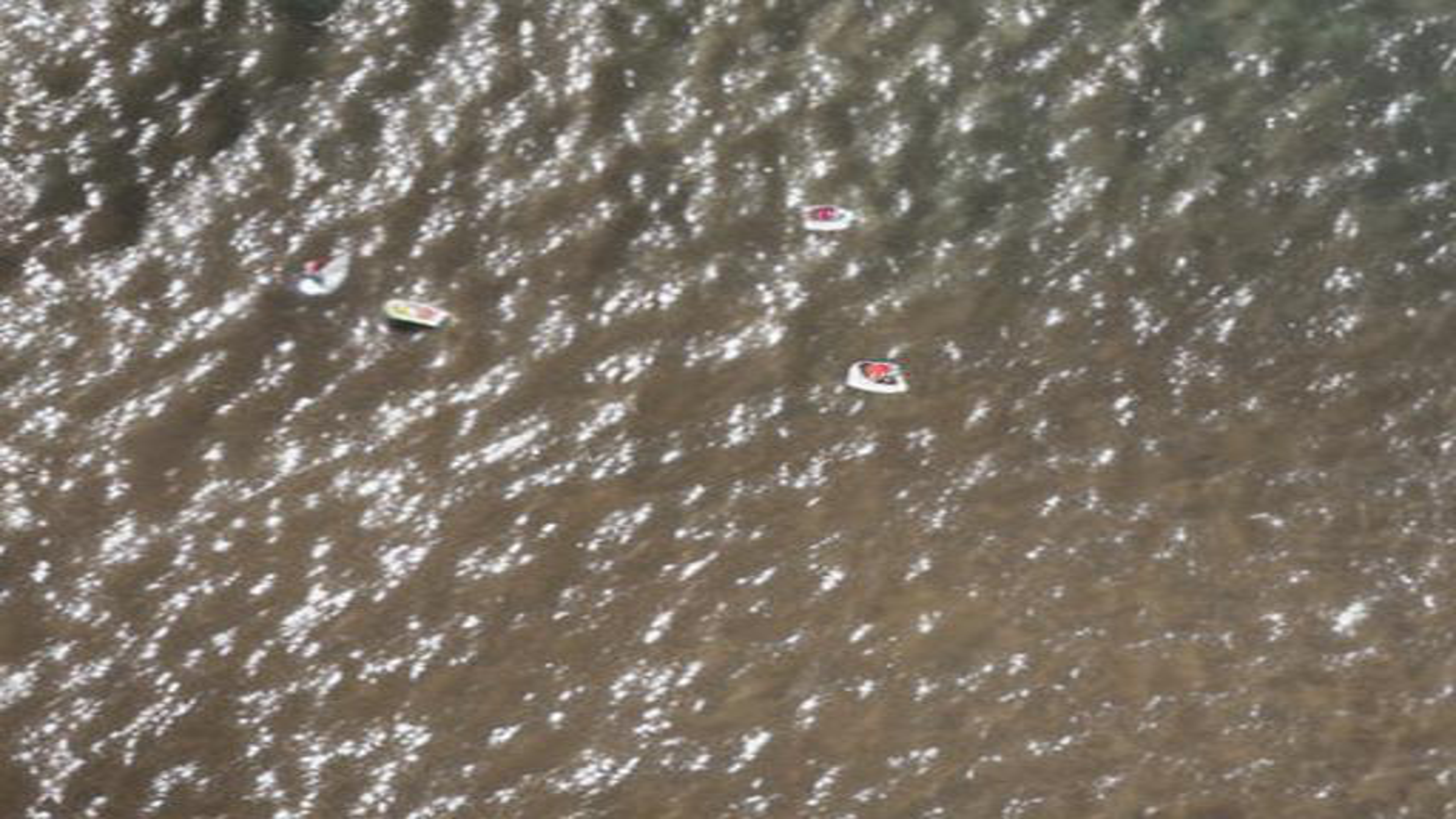}
    \caption*{Edited}
  \end{minipage}
\caption{Example of good edited images using Blended Latent Diffusion. The image's sea background is edited while the objects are preserved.}
\label{fig:BLD}
\end{figure}

 When employing Blended Latent Diffusion, the issue of objects of interest being uncontrollably spread is mitigated, as demonstrated in Figure \ref{fig:BLD}. Notably, the boat objects are effectively preserved, and the edited background exhibits acceptable quality. Nevertheless, it is worth mentioning that the generated sea background does not possess the same vividness observed in Stable Diffusion Inpainting. This disparity may be affected by the utilisation of different trained Diffusion models in these approaches. Future work will investigate other image generation methods that have similar preservation properties to the Blended Latent Diffusion whilst producing more vivid background. 
 
 

To further confirm the above result, we present the percentage of images produced by SD Inpainting and BLD retained after applying our SafeSea filter. Results in Table~\ref{table:comparison} suggest that SD Inpainting produces much lower quality than BLD as significant amount of images are filtered.


\begin{table}[h!]
\centering
\begin{tabular}{@{}lcccc@{}}
    \toprule
     & SS1 & SS2 & SS3 & SS4 \\ 
     \midrule
     Generated & 2,336 & 25,114 & 65,275 & 4,275 \\ 
     Filtered & 2,087 & 19,390 & 45,066 & 3,151 \\  
     \bottomrule
\end{tabular}
     \caption{Number of images generated before and after applying object preservation checker. 97,000 images are generated from 300 sourced images and then filter out the ones that do not retain any object.}
     \label{table:1}
\end{table}

\begin{table}[!h]
\centering
\begin{tabular}{@{}lcc@{}}
    \toprule
     & SD Inpainting~\cite{stableDiffusion}  & BLD~\cite{BlendedLatentDiffusion} \\ 
     \midrule
     Image passing rate (in \%)   & 8.16\% & \textbf{71.85\%} \\ 
     \bottomrule
\end{tabular}
 \caption{Comparison between SD Inpainting and BLD when we apply our SafeSea filters. The Image passing rate computes the percentage of images passed by our filters. High passing rate suggests better quality images according to our filters.}
 \label{table:comparison}
 
\end{table}
\vspace{-5mm}
\subsection{Applying YoloV5 on the SafeSea dataset}
We employed the SafeSea dataset for assessing a pre-trained YoloV5 object detection model's ability in detecting 'boat' objects across various Sea State levels. 

\subsubsection{Experiment setup}

We run the YoloV5 with default parameters on the SafeSea dataset images. the Mean Average Precision (mAP) for the four sea state level is then calculated.
\vspace{-3mm}
\subsubsection{Results}
\begin{figure}[b]
    \centering
    \includegraphics[width=\linewidth]{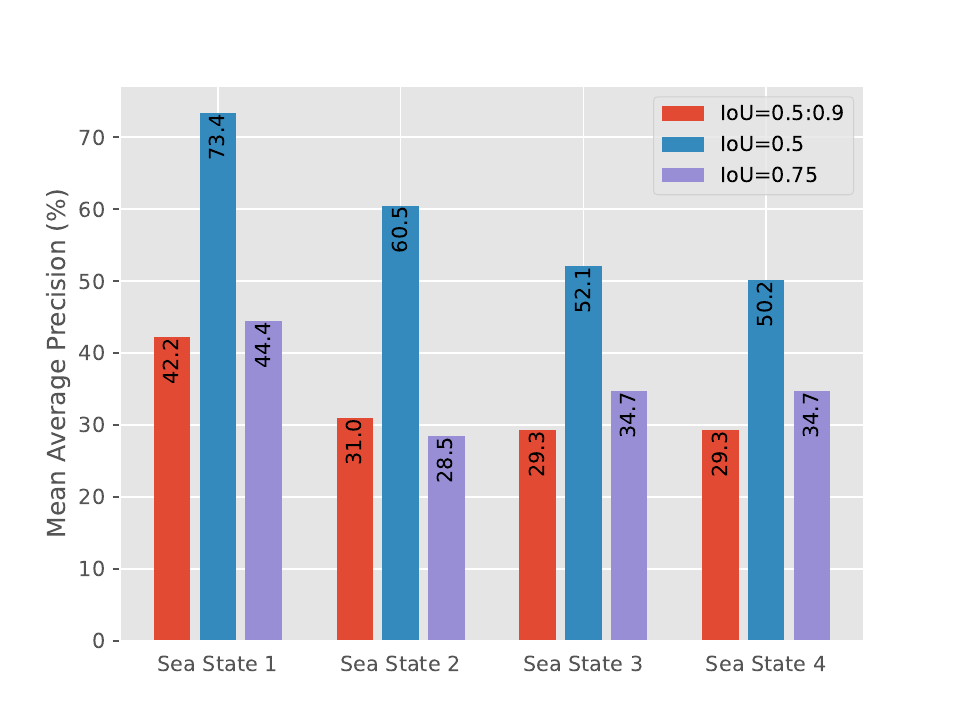}
    \caption{mAP Comparison for Different Sea State Levels. It shows that the mAP values decrease from Sea State 1 to Sea State 4 for both IoU of 0.5 and from 0.5 to 0.95.}
    \vspace{-4mm}
    \label{fig:mAP}
\end{figure}

\vspace{-2mm}
Figure \ref{fig:mAP} presents the result. Notably, there is a noticeable decrease in mAP values from Sea State 1 to Sea State 4, both for IoU of 0.5 and the range of 0.5 to 0.95.

Given that the Sea State Level background is not entirely controlled within the transformation process, the distribution of images among the sea state categories is determined by the Sea State Classifier. The results indicate a significantly higher number of images classified as Sea State 3 of 45,066; while Sea State 1 comprises the lowest number with 2,087 images. It is important to note that each image originally contains a varying number of objects, both before and after the editing process. Additionally, object size plays a considerable role in the editing process; we have observed that larger objects tend to be better preserved in comparison to smaller ones. This, in turn, has a direct impact on the object detection confidence scores, which subsequently affect the mAP scores. In general, we observe that the pre-trained model tends to be more struggle when detecting objects in images classified with higher sea state levels. However, it is crucial to acknowledge that several other factors also influence the results, such as the number of objects being preserved and the sizes of the original objects.


\vspace{-3mm}
\section{Conclusion}
\label{sec:conclusion}
In this work, we introduce SafeSea, a proof-of-concept for generating synthetic maritime images by modifying real images, where the original sea background is transformed to simulate various sea surface conditions, corresponding to different Sea States. Our method capitalizes on the capabilities of Blended Latent Diffusion \cite{BlendedLatentDiffusion} to manipulate images. Subsequently, these modified images are categorized into distinct Sea State Levels, ranging from 1 to 4. Moreover, the original objects within these images are scrutinized to ensure their preservation throughout the editing process. Employing this technique, we have created the SafeSea dataset, which includes maritime images featuring marine objects set against diverse Sea State backgrounds by utilizing the 'SeaDroneSea' dataset~\cite{varga2022seadronessee}. Additionally, we have observed that stormy sea backgrounds can impact the performance of the YoloV5 object detection model.

\vspace{-2mm}
\section{Limitation and Future Work}
\vspace{-1mm}
\label{sec:futurework}
The current SeaSafe method, as proposed, exhibits certain limitations that necessitate attention in our future work. Primarily, it lacks control over the generated sea background during the editing process, limiting the diversity of realistic backgrounds. Furthermore, the diffusion model employed by BLD \cite{BlendedLatentDiffusion} has constraints in generating realistic wave and whitecap patterns. Additionally, smaller objects such as swimmers are ignored in the image quality evaluation. Our forthcoming efforts will concentrate on optimizing the image editing process to elevate the overall quality of generated images. Exploring alternative image editing methods is also on the agenda to enhance the image generation module. Additionally, addressing the control over the insertion of irrelevant objects during editing is crucial, as it can significantly impact object detection models. Future work will specifically tackle the introduction of unexpected objects, mitigating their potential impact on object detection models. Simultaneously, improvements are planned for both the Sea State Classifier and Object Detection Checker to elevate their performance. Furthermore, we aim to implement an additional filter to exclude generated images that do not align with the desired Sea State Level criteria. These enhancements collectively constitute our roadmap for refining the SeaSafe method in subsequent stages of development. Future work will also delve into the scalability of SafeSea on larger datasets and in real-world scenarios beyond the SeaDronesSee dataset with the exploration of how the object detector performs against the unedited SeaDronesSee dataset.
\vspace{-2mm}
\section{Acknowledgement}
\vspace{-1mm}
\label{sec:acknowledgement}
This work has been supported by Sentient Vision Systems. Sentient Vision Systems is one of the leading Australian developers of computer vision and artificial intelligence software solutions for defence and civilian applications.
  
{
    \small
    \bibliographystyle{ieee_fullname}
    \bibliography{main}
}


\end{document}